\documentclass[11pt]{article}
\usepackage{eacl2017}
\usepackage{times}
\usepackage{textpos}
\usepackage{url}
\usepackage{latexsym}
\usepackage{tikz}
\usepackage{hhline}
\usetikzlibrary{shapes,snakes}
\usepackage{amsmath}
\usepackage{multirow}
\usepackage{graphicx}
\usepackage{listings}
\usepackage{amsthm}
\usepackage{xcolor}
\usepackage{comment}

\eaclfinalcopy 
\def\eaclpaperid{339} 

\theoremstyle{definition}

\title{An Incremental Parser for Abstract Meaning Representation}

\author{
Marco Damonte \\
School of Informatics \\ University of Edinburgh \\
{\small \texttt{m.damonte@sms.ed.ac.uk}} \\\And
Shay B. Cohen \\
School of Informatics \\ University of Edinburgh \\
{\small \texttt{scohen@inf.ed.ac.uk}} \\\And
Giorgio Satta \\
Dept. of Information Engineering \\ University of Padua \\
{\small \texttt{satta@dei.unipd.it}}
}

\date{}

\newcommand{\arceager}{{\sc ArcEager}}
\newcommand{\arcstandard}{{\sc ArcStandard}}
\newcommand{\amreager}{{\sc AmrEager}}
\newcommand{\conceptfunc}{\pi}

\newcommand{\troot}{\circ}
\newcommand{\tshift}{{\sf Shift}}
\newcommand{\tlarc}{{\sf LArc}}
\newcommand{\trarc}{{\sf RArc}}
\newcommand{\treduce}{{\sf Reduce}}
\newcommand{\sep}{\; | \;}
\newcommand{\termdef}[1]{\textbf{{#1}}}
\newcommand{\sset}[1]{{\cal S}({#1})}

\newcommand{\order}[1]{{\cal O}({#1})}

\newcommand{\ignore}[1]{}

\includecomment{arxiv}\excludecomment{eacl}
\begin{arxiv}
\end{arxiv}

\begin{document}

\maketitle

\begin{arxiv}
\begin{textblock}{10}(0,-4.1)
{\small
\noindent\textcolor{gray}{From the Proceedings of EACL 2017 (Valencia, Spain). This version includes slightly more information than the published version (January, 2017).}
}
\end{textblock}
\end{arxiv}

\begin{abstract}
Abstract Meaning Representation (AMR) is a semantic representation 
for natural language that 
embeds annotations related to traditional tasks
such as named entity recognition, semantic role labeling, 
word sense disambiguation and co-reference resolution.
We describe a transition-based parser for AMR 
that parses sentences left-to-right, in linear time. 
We further propose a test-suite that assesses 
specific subtasks that are helpful in comparing AMR parsers, and 
show that our parser is competitive with the state 
of the art on the LDC2015E86 dataset and that it outperforms 
state-of-the-art parsers for recovering named entities and handling polarity.
\end{abstract}

\section{Introduction}
Semantic parsing aims to solve the problem of canonicalizing
language and representing its meaning: given an input sentence, it aims to extract a semantic representation of that sentence. Abstract meaning representation \cite{Banarescu13abstractmeaning}, or AMR for short, allows us to do that with the inclusion of most of the shallow-semantic natural language processing (NLP) tasks that are
usually addressed separately, such as named entity recognition, 
semantic role labeling 
and co-reference resolution.
AMR is partially motivated by the need to provide the NLP  community with a single dataset that includes basic disambiguation information, instead of having to rely on different datasets for each disambiguation problem. The annotation process is straightforward, enabling the development of large datasets.
\begin{eacl}
Alternative semantic representations have been developed and studied, such as CCG \cite{surface_steedman,syntax_steedman} and UCCA \cite{abend2013universal}.
\end{eacl}

Several parsers for AMR have been recently developed \cite{carbonell2014discriminative,wang2boosting,peng2015synchronous,pust2015using,goodman2016noise,searn_amr,microsoft,artzi2009broad,barzdins2016riga,emnlp2016}.  This line of research is new and current results suggest a large room for improvement. Greedy transition-based methods~\cite{Nivre-J08-4003} are one of the most popular choices for dependency parsing, because of their good balance between efficiency and accuracy. These methods seem promising also for AMR, due to the similarity between dependency trees and AMR structures, i.e., both representations use graphs with nodes that have lexical content and edges that represent linguistic relations.

A transition system is an abstract machine characterized by a set of configurations and transitions between them. The basic components of a configuration are a stack of partially processed words and a buffer of unseen input words.  Starting from an initial configuration, the system applies transitions until a terminal configuration is reached. The sentence is scanned left to right, with linear time complexity for dependency parsing. This is made possible by the use of a greedy classifier that chooses the transition to be applied at each step. 



In this paper we introduce a parser for AMR that is inspired by the \arceager{} dependency transition system of \newcite{nivre2004}. 
The main difference between our system and \arceager{} is that we need to account for the mapping from word tokens to AMR nodes, non-projectivity of AMR structures and reentrant nodes (multiple incoming edges). Our AMR parser brings closer dependency parsing and AMR parsing by showing that dependency parsing algorithms, with some modifications, can be used for AMR. Key properties such as working left-to-right, incrementality\footnote{Strictly speaking, transition-based parsing cannot achieve full incrementality, which requires to have a single connected component at all times \cite{nivre2004}.} and linear complexity further strengthen its relevance.

The AMR parser of \newcite{wang2boosting}, called CAMR, also defines a transition system. It differs from ours because we process the sentence left-to-right while they first acquire the entire dependency tree and then process it bottom-up. 
More recently \newcite{emnlp2016} presented a non-greedy transition system for AMR parsing, based on \arcstandard{} \cite{nivre2004}. Our transition system is also related to an adaptation of \arceager{} for directed acyclic graphs (DAGs), introduced by \newcite{sagae2008shift}. This is also the basis for \newcite{ribeyre2015because}, a transition system used to parse dependency graphs. Similarly, \newcite{du2014peking} also address dependency graph parsing by means of transition systems. Analogously to dependency trees, dependency graphs have the property that their nodes consist of the word tokens, which is not true for AMR. As such, these transition systems are more closely related to traditional transition systems for dependency parsing.

Our contributions in this paper are as follows:
\begin{itemize}
\item
In \S\ref{se & ts} we develop a left-to-right, linear-time transition system for AMR parsing, inspired by the \arceager{} transition system for dependency tree parsing;
\item
In \S\ref{se & metr} we claim that the Smatch score \cite{cai2013smatch} is not sufficient to evaluate AMR parsers and propose a set of metrics to alleviate this problem and better compare alternative parsers;

\item
In \S\ref{se & exp} we show that our algorithm is competitive with publicly available state-of-the-art parsers on several metrics.
\end{itemize}

\section{Background and Notation}
\label{se & reent}

\paragraph{AMR Structures}

AMRs are rooted and directed graphs with node and edge labels. An annotation example for the sentence \emph{I beg you to excuse me} is shown in Figure~\ref{fig:ann}, with the AMR graph reported in Figure~\ref{fig:beg}.

\begin{figure}
\begin{lstlisting}[frame=single]
((*@\textbf{b}@*) / (*@\textit{beg-01}@*)
    (*@\textit{:ARG0}@*) ((*@\textbf{i}@*) / (*@\textit{i}@*)
    (*@\textit{:ARG1}@*) ((*@\textbf{y}@*) / (*@\textit{you)}@*)
    (*@\textit{:ARG2}@*) ((*@\textbf{e}@*) / (*@\textit{excuse-01}@*)
	(*@\textit{:ARG0}@*) (*@\textbf{y}@*)
	(*@\textit{:ARG1}@*) (*@\textbf{i}@*)))
\end{lstlisting}
\caption{Annotation for the sentence ``\emph{I beg you to excuse me}.'' Variables are in boldface and concepts and edge labels are in italics.}
\label{fig:ann}
\end{figure}

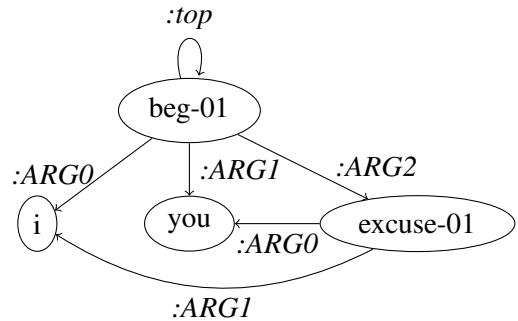
\begin{figure}[t]
 \centering
  \begin{tikzpicture}

        \draw (0,10) node(beg)[ellipse,draw] {beg-01};
        \draw (-2,8.5) node(i)[ellipse,draw] {i};
        \draw (0,8.5) node(you)[ellipse,draw] {you};
        \draw (3,8.5) node(excuse)[ellipse,draw] {excuse-01};

        \draw [->] (beg) -- node[left]{\emph{:ARG0}} (i);
        \draw [->] (beg) -- node[right]{\emph{:ARG1}} (you);
        \draw [->] (beg) -- node[right=0.3cm]{\emph{:ARG2}} (excuse);
        \draw[->]  (excuse) to node [auto] {\emph{:ARG0}} (you);
        \draw[bend left,->]  (excuse) to node [auto] {\emph{:ARG1}} (i);
        \draw [->] (beg) edge[loop above]node{\emph{:top}} (beg);
  \end{tikzpicture}

 \caption{AMR graph representation for Figure~\ref{fig:ann}.}
 \label{fig:beg}
\end{figure}



Concepts are represented as labeled nodes in the graph and can be either English words (e.g. \emph{I} and
\emph{you}) or Propbank framesets (e.g. \emph{beg-01} and \emph{excuse-01}). Each node in the graph
is assigned to a variable in the AMR annotation so that a variable re-used in the annotation corresponds to reentrancies
(multiple incoming edges) in the graph. Relations are represented as labeled and directed edges in
the graph.

\paragraph{Notation}

For most sentences in our dataset, the AMR graph is a directed acyclic graph (DAG), with a few specific cases where cycles are permitted. These cases are rare, and for the purpose of this paper, we consider AMR as DAGs.


We denote by $[n]$ the set $\{ 1, \ldots, n \}$. We define an AMR structure as a tuple $(G, x, \conceptfunc)$, where $x = x_1 \cdots x_n$ is a sentence, with each $x_i$, $i \in [n]$, a word token, and $G$
is a directed graph $G=(V,E)$ with $V$ and $E$ the set of nodes and edges, respectively.\footnote{We collapse all multi-word named entities in a single token (e.g., \emph{United Kingdom} becomes \emph{United\_Kingdom}) both in training and parsing.} We assume $G$ comes along with a node labeling function and an edge labeling function.
Finally, $\conceptfunc \colon V \rightarrow [n]$ is a total alignment function that maps every node of the graph to an index $i$ for the sentence $x$, with the meaning that node $v$ represents (part of) the concept expressed by the word $x_{\conceptfunc(v)}$.\footnote{$\conceptfunc$ is a function because we do not consider co-references, which would otherwise cause a node to map to multiple indices. This is in line with current work on AMR parsing.}

We note that the function $\conceptfunc$ is not invertible, since it is neither injective nor surjective.  For each $i \in [n]$, we let 
\begin{equation*}
\conceptfunc^{-1}(i) = \{ v \sep v \in V, \; \conceptfunc(v) = i\}
\end{equation*}
be the pre-image of $i$ under $\conceptfunc$ (this set can be empty for some $i$), which means that we map a token in the sentence to a set of nodes in the AMR. In this way we can align each index $i$ for $x$ to the induced subgraph of $G$.  More formally, we define
\begin{equation}\label{eq:subgraph}
\overleftarrow{\conceptfunc}(i) = ( \conceptfunc^{-1}(i), E \cap (\conceptfunc^{-1}(i) \times \conceptfunc^{-1}(i)) ),
\end{equation}
with the node and edge labeling functions of $\overleftarrow{\conceptfunc}(i)$ inherited from $G$. Hence, $\overleftarrow{\conceptfunc}(i)$ returns the AMR subgraph aligned with a particular token in the sentence.


\subsection{Transition-Based AMR Parsing}
\label{se & tsdp}

Similarly to dependency parsing, AMR parsing is partially based on the identification of predicate-argument structures. Much of the dependency parsing literature focuses on {\em transition-based} dependency parsing---an approach to parsing that scans the sentence from left to right in linear time and updates an intermediate structure that eventually ends up being a dependency tree.

\begin{arxiv}
The two most common transition systems for greedy dependency parsing are \arcstandard{} and \arceager{}. With \arcstandard{}, a stack is maintained along with a buffer on which the left-to-right scan is performed. At each step, the parser chooses to scan a word in the buffer and shift it onto the stack, or else to create an arc between the two top-most elements in the stack and pop the dependent.  \arcstandard{} parses a sentence in a pure bottom-up, left-to-right fashion (similarly to shift-reduce context-free grammar parsers), and must delay the construction of right arcs until all the dependent node has been completed.  This imposes strong limitations on the degree of incrementality of the parser.
The \arceager{} system was designed to improve on \arcstandard{} by mixing bottom up and top-down strategies.  More precisely, in the \arceager{} parser left arcs are constructed bottom-up and  right arcs are constructed top-down, so that right dependents can be attached to their heads even if some of their own dependents are not identified yet.  In this way arcs are constructed as soon as the head and the dependent are available in the stack.
\end{arxiv}

Because of the similarity of AMR structures to dependency structures, transition systems are also helpful for AMR parsing.  Starting from the \arceager{} system, we develop here a novel transition system, called \amreager{} that parses sentences into AMR structures. There are three key differences between AMRs and dependency trees that require further adjustments for dependency parsers to be used with AMRs.

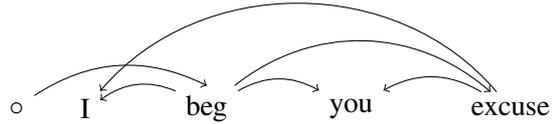
\begin{figure}[t]
 \centering
  \begin{tikzpicture}
	\fill (0,10) node(root) {$\troot{}$};
	\draw (0.9,10) node(i) {I};
        \draw (2.5,10) node(beg) {beg};
        \draw (4.4,10) node(you) {you};
        \draw (6.5,10) node(excuse) {excuse};

         \draw [bend right,->] (beg) to node[right]{} (i);
         \draw [bend left,->] (beg) to node[left]{} (you);
         \draw [bend right,->] (excuse) to node [right] {} (you);
         \draw [bend right=50,->] (excuse) to node [right] {} (i);
         \draw [bend left=40,->] (beg) to node[left]{} (excuse);
         \draw [bend left,->] (root) to node[left]{} (beg.north);
  \end{tikzpicture}

 \caption{AMR's edges for the sentence ``I beg you to excuse me.'' mapped back to the sentence, according to the alignment. $\troot{}$ is a special token representing the root.}
 \label{fig:beg_dep}
\end{figure}

\paragraph{Non-Projectivity}
A key difference between English dependency trees and AMR structures is projectivity. Dependency trees in English are usually projective, roughly meaning that there are no crossing arcs if the edges are drawn in the semi-plane above the words. While this restriction is empirically motivated in syntactic theories for English, it is no longer motivated for AMR structures.

The notion of projectivity can be generalized to AMR graphs as follows.  The intuition is that we can use the alignment $\conceptfunc$ to map AMR edges back to the sentence $x$, and test whether there exist pairs of crossing edges. Figure~\ref{fig:beg_dep} shows this mapping for the AMR of Figure~\ref{fig:beg}, where 
the edge connecting \emph{excuse} to \emph{I} crosses another edge.
More formally, consider an AMR edge $e = (u,\ell,v)$.  Let $\conceptfunc(u) = i$ and $\conceptfunc(v) = j$, so that $u$ is aligned with $x_{i}$ and $v$ is aligned with $x_{j}$.  The spanning set for $e$, written $\sset{e}$, is the set of all nodes $w$ such that $\conceptfunc(w) = k$ and $i < k < j$ if $i < j$ or $j < k < i$ if $j < i$.  We say that $e$ is \termdef{projective} if, for every node $w \in \sset{e}$, all of its parent and child nodes are in $\sset{e} \cup \{u,v\}$; otherwise, we say that $e$ is \termdef{non-projective}.  An AMR is projective if all of its edges are projective, and is non-projective otherwise. This corresponds to the intuitive definition of projectivity for DAGs introduced in \newcite{sagae2008shift} and is closely related to the definition of non-crossing graphs of \newcite{kuhlmann2015parsing}.

Table~\ref{tab:stats} demonstrates that a relatively small percentage of all AMR edges are non-projective. Yet, 35\% of the sentences contain at least one
non-projective edge.


\begin{table}
  \begin{center}
  \begin{tabular}{|l|c|}
    \hline
    Non-projective edges & 8\%\\
    Non-projective AMRs & 35\%\\
    Reentrant edges & 7\%\\
    AMRs with at least one reentrancy & 51\%\\
    \hline
  \end{tabular}
  \end{center}
  \caption{Statistics for non-projectivity and reentrancies in 200 AMR manually aligned with the associated sentences.\protect\footnotemark}
  \label{tab:stats}
\end{table}
\footnotetext{\url{https://github.com/jflanigan/jamr/blob/master/docs/Hand_Alignments.md}}

\paragraph{Reentrancy} AMRs are graphs rather than trees because they can have nodes with multiple parents, called reentrant nodes, as in the node \emph{you} for the AMR of Figure~\ref{fig:beg}. There are two phenomena that cause reentrancies in AMR: control, where a reentrant edge appears between siblings of a control verb, and co-reference, where multiple mentions correspond to the same concept.\footnote{A valid criticism of AMR is that these two reentrancies are of a completely different type, and should not be collapsed together. Co-reference is a discourse feature, working by extra-semantic mechanisms and able to cross sentence boundaries, which are not crossed in AMR annotation.}

In contrast, dependency trees do not have nodes with multiple parents.  Therefore, when creating a new arc, transition systems for dependency parsing check that the dependent does not already have a head node, preventing the node from having additional parents.
To handle reentrancy, which is not uncommon in AMR structures as shown in Table~\ref{tab:stats}, we drop this constraint.


\paragraph{Alignment}
Another main difference with dependency parsing is that in AMR there is no straightforward mapping
between a word in the sentence and a node in the graph: words may generate no nodes, one node or multiple
nodes. In addition, the labels at the nodes are often not easily determined by the word in the
sentence. For instance \emph{expectation} translates to \emph{expect-01} and \emph{teacher} translates to the two nodes \emph{teach-01} and \emph{person}, connected through an \emph{:ARG0} edge, expressing that a teacher is a person who teaches. A mechanism of concept identification is therefore required to map each token $x_i$ to a subgraph with the correct labels at its nodes and edges: if $\conceptfunc$ is the gold alignment, this should be the subgraph $\overleftarrow{\conceptfunc}(i)$ defined in Equation~(\ref{eq:subgraph}). To obtain alignments between the tokens in the sentence and the nodes in the AMR graph of our training data, we run the JAMR aligner.\footnote{\url{https://github.com/jflanigan/jamr}}

\section{Transition system for AMR Parsing}
\label{se & ts}

A \termdef{stack} $\sigma  = \sigma_n | \cdots | \sigma_1 | \sigma_0$ is a list of nodes of the partially constructed AMR graph, with the top element $\sigma_0$ at the right.  We use the symbol `$|$' as the concatenation operator. A \termdef{buffer} $\beta = \beta_0 | \beta_1 | \cdots | \beta_n $ is a list of indices from $x$, with the first element $\beta_0$ at the left, representing the word tokens from the input still to be processed. A \termdef{configuration} of our parser is a triple $(\sigma, \beta, A)$, where $A$ is the set of AMR edges that have been constructed up to this point. 

In order to introduce the transition actions of our parser we need some additional notation.
We use a function $a$ that maps indices from $x$ to AMR graph fragments.  For each $i \in [n]$, $a(i)$ is a graph $G_a = (V_a, E_a)$, with single root $\mathrm{root}(G_a)$, representing the semantic contribution of word $x_i$ to the AMR for $x$.  As already mentioned, $G_a$ can have a single node representing the concept associated with $x_i$, or it can have several nodes in case $x_i$ denotes a complex concept, or it can be empty.



The transition \tshift{} is used to decide if and what to push on the stack after consuming a token from the buffer.  Intuitively, the graph fragment $a(\beta_0)$ obtained from the token $\beta_0$, if not empty, is ``merged'' with the graph we have constructed so far. We then push onto the stack the node $\mathrm{root}(a(\beta_0))$ for further processing.
\tlarc{}$(\ell)$ creates an edge with label $\ell$ between the top-most node and the second top-most node in the stack, and pops the latter.
\trarc{}$(\ell)$ is the symmetric operation, but does not pop any node from the stack. 

Finally, \treduce{} pops the top-most node from the stack, and it also recovers reentrant edges between its sibling nodes, capturing for instance several control verb patterns. To accomplish this, \treduce{} decides whether to create an additional edge between the node being removed and the previously created sibling in the partial graph.  
This way of handling control verbs is similar to the \emph{REENTRANCE} transition of \newcite{wang2boosting}. 

The choice of popping the dependent in the \tlarc{} transition is inspired by \arceager{}, where left-arcs are constructed bottom-up to increase the incrementality of the transition system \cite{nivre2004}. This affects our ability to recover some reentrant edges: consider a node $u$ with two parents $v$ and $v'$, where the arc $v \rightarrow u$ is a left-arc and $v' \rightarrow  u$ is any arc. If the first arc to be processed is $v \rightarrow u$, we use \tlarc{} that pops $u$, hence making it impossible to create the second arc $v' \rightarrow u$. Nevertheless, we discovered that this approach works better
than a completely unrestricted allowance of reentrancy. The reason is that if we do not remove dependents at all when first attached to a node, the
stack becomes larger, and nodes which should be connected end up being distant from each other, and as such, are never connected. 

The initial configuration of the
system has a $\troot$ node (representing the root) in the stack and the entire sentence in the buffer.
The terminal configuration consists of an empty buffer and a stack with only the $\troot$ node.  
The transitions required to parse
the sentence \emph{The boy and the girl} are shown in Table~\ref{tab:bg}, where the first line shows the initial configuration and the last line shows the terminal configuration.

Similarly to the transitions of the \arceager{}, the above transitions construct edges as soon as the head and the dependent are available in the stack, with the aim of maximizing the parser incrementality. 
We now show that our greedy transition-based AMR parser is linear-time in $n$, the length of the input sentence $x$.  We first claim that the output graph has size $\order{n}$.  Each token in $x$ is mapped to a constant number of nodes in the graph by \tshift{}.  Thus the number of nodes is $\order{n}$.  Furthermore, each node can have at most three parent nodes, created by transitions \trarc{}, \tlarc{} and \treduce{}, respectively.  Thus the number of edges is also $\order{n}$.  It is possible to bound the maximum number of transitions required to parse $x$: the number of \tshift{} is bounded by $n$, and the number of \treduce{}, \tlarc{} and \trarc{} is bounded by the size of the graph, which is $\order{n}$. Since each transition can be carried out in constant time, we conclude that our parser runs in linear time. 



\begin{table*}[t]
\begin{center}
  \begin{tabular}{|l|l|l|l|}
    \hline
    action & stack & buffer & edges \\
    \hline
    - & $\lbrack \troot \rbrack$ & $\lbrack$the,boy,and,the,girl$\rbrack$ & $\{\}$ \\
    \tshift{} & $\lbrack \troot \rbrack$ & $\lbrack$boy,and,the,girl$\rbrack$ & $\{\}$ \\
    \tshift{} & $\lbrack \troot$, \emph{boy}$ \rbrack$ & $\lbrack$and,the,girl$\rbrack$ & $\{\}$ \\
    \tshift{} & $\lbrack \troot$, \emph{boy}, \emph{and} $\rbrack$ & $\lbrack$the,girl$\rbrack$ & $\{\}$ \\
    \tlarc{} & $\lbrack \troot$, \emph{and} $\rbrack$ & $\lbrack$the,girl$\rbrack$ & $\{\langle$\emph{and,:op1,boy}$\rangle\} = A_1$ \\
    \trarc{} & $\lbrack \troot$, \emph{and} $\rbrack$ & $\lbrack$the,girl$\rbrack$ &
    $A_1 \cup \{\langle\troot$\emph{,:top,and}$\rangle\} = A_2$ \\
    \tshift{} & $\lbrack \troot$, \emph{and} $\rbrack$ & $\lbrack$girl$\rbrack$ & $A_2$ \\
    \tshift{} & $\lbrack \troot$, \emph{and}, \emph{girl} $\rbrack$ & $\lbrack\rbrack$ & $A_2$ \\
    \trarc{} & $\lbrack \troot$, \emph{and}, \emph{girl} $\rbrack$ & $\lbrack\rbrack$ & $A_2 \cup \{\langle$\emph{and,:op2,girl}$\rangle\} = A_3$ \\
    \treduce{} & $\lbrack \troot$, \emph{and} $\rbrack$ & $\lbrack\rbrack$ & $A_3$ \\
    \treduce{} & $\lbrack \troot \rbrack$ & $\lbrack\rbrack$ & $A_3$ \\
    \hline
  \end{tabular}
\end{center}
  \caption{Parsing steps for the sentence ``\emph{The boy and the girl}.''}
  \label{tab:bg}
\end{table*}




\section{Training the System}
\label{se & train}

Several components have to be learned: (1) a transition classifier that predicts the next transition given the current configuration, (2) a binary classifier that decides whether or not to create a reentrancy after a \treduce{}, (3) a concept identification step for each \tshift{} to compute $a(\beta_0)$, and 3) another classifier to label edges after each \tlarc{} or \trarc{}.

\subsection{Oracle}
\label{se & oracle}

Training our system from data requires an oracle---an algorithm that given a gold-standard AMR graph and a sentence returns transition sequences that maximize the overlap between the gold-standard graph and the graph dictated by the sequence of transitions.

We adopt a shortest stack, static oracle similar to \newcite{manningfast}.  Informally, static means that if the actual configuration of the parser has no mistakes, the oracle provides a transition that does not introduce any mistake.  Shortest stack means that the oracle prefers transitions where the number of items in the stack is minimized. Given the current configuration $(\sigma, \beta, A)$ and the gold-standard graph $G = (V_g,A_g)$, the oracle is defined as follows, where we test the conditions in the given order and apply the action associated with the first match:

\begin{enumerate}
\item
if $\exists \ell[(\sigma_0,\ell,\sigma_1) \in A_g]$ then \tlarc{}($\ell$);
\item
if $\exists \ell[(\sigma_1,\ell,\sigma_0) \in A_g]$ then \trarc{}($\ell$);
\item
if $\neg \exists i,\ell[(\sigma_0,\ell,\beta_i) \in A_g \vee (\beta_i,\ell,\sigma_0) \in A_g]$ then \treduce{};
\item
\tshift{} otherwise.
\end{enumerate}

The oracle first checks whether some gold-standard edge can be constructed from the two elements at the top of the stack (conditions 1 and 2).  If \tlarc{} or \trarc{} are not possible, the oracle checks whether all possible edges in the gold graph involving $\sigma_0$ have already been processed, in which case it chooses \treduce{} (conditions 3).  To this end, it suffices to check the buffer, since \tlarc{} and \trarc{} have already been excluded and elements in the stack deeper than position two can no longer be accessed by the parser.  If \treduce{} is not possible, \tshift{} is chosen.

Besides deciding on the next transition, the oracle also needs the alignments, which we generate with JAMR, in order to know how to map the next token in the sentence to its AMR subgraph $\overleftarrow{\conceptfunc}(i)$ defined in~(\ref{eq:subgraph}).


\subsection{Transition Classifier}
\begin{table}[t]
  \begin{tabular}{|l|l|}
    \hline
    depth & $d(\sigma_0), d(\sigma_1)$\\
    children & $\#c(\sigma_0), \#c(\sigma_1)$ \\
    parents & $\#p(\sigma_0), \#p(\sigma_1)$\\
    lexical & $w(\sigma_0), w(\sigma_1), w(\beta_0), w(\beta_1)$, \\
	& $w(p(\sigma_0)), w(c(\sigma_0)), w(cc(\sigma_0))$, \\
	& $w(p(\sigma_1)), w(c(\sigma_1)), w(cc(\sigma_1))$\\
    POS & $s(\sigma_0), s(\sigma_1), s(\beta_0), s(\beta_1)$\\
    entities & $e(\sigma_0), e(\sigma_1), e(\beta_0), e(\beta_1)$ \\
    dependency & $\ell(\sigma_0,\sigma_1), \ell(\sigma_1, \sigma_0)$,\\
    & $\forall i \in \{ 0, 1\}$: $\ell(\sigma_i,\beta_0), \ell(\beta_0,\sigma_i)$\\
      & $\forall i \in \{ 1, 2, 3\}$: $\ell(\beta_0,\beta_i), \ell(\beta_i,\beta_0)$\\
      & $\forall i \in \{ 1, 2, 3\}$: $\ell(\sigma_0,\beta_i), \ell(\beta_i,\sigma_0)$\\
    \hline
  \end{tabular}
  \caption{Features used in transition classifier.
The function $d$ maps a stack element to the depth of the associated graph fragment.
The functions $\#c$ and $\#p$ count the number of children and parents, respectively, of a stack element. The function $w$ maps a stack/buffer element to the word embedding for the associated word in the sentence. The function $p$ gives the leftmost (according to the alignment) parent of a stack element, the function $c$ the leftmost child and the function $cc$ the leftmost grandchild. The function $s$ maps a stack/buffer element to the part-of-speech embedding for the associated word. The function
$e$ maps a stack/buffer element to its entity. Finally, the function $\ell$ maps a pair of symbols to the dependency label embedding, according to the edge (or lack of) in the dependency tree for the two words these symbols are mapped to.}
  \label{tab:features}
\end{table}

Like all other transition systems of this kind, our transition system has a ``controller'' that predicts a transition given the current configuration (among \tshift{}, \tlarc{}, \trarc{} and \treduce{}).
The examples from which we learn this controller are based on features extracted from the oracle transition
sequences, where the oracle is applied on the training data.

\ignore{
The job of the transition classifier is to decide which transition (among \tshift{}, \tlarc{}, \trarc{} and \treduce{}) to
choose in a given state. We learn this in a supervised fashion by creating a dataset of states and
gold actions for those states, which is generated by parsing each sentence with the oracle and
recording the action chosen by the oracle at each stat &
\begin{equation}
D = \{c_i, oracle(c_i)\}
\end{equation}
The alignments required by the oracle are automatically generated with JAMR, as the
200 sentences manually aligned are too few to train the classifier.
}

As a classifier, we use a feed-forward neural network with two hidden layers of 200 tanh units and learning rate set to 0.1, with linear decaying.
The input to the network consists of the concatenation of embeddings for words, POS tags and Stanford parser
dependencies, one-hot vectors for named entities and additional sparse features, extracted from the current configuration of the transition system; this is reported in more details in Table~\ref{tab:features}.
The embeddings for words and POS tags were
pre-trained on a large unannotated corpus consisting of the first 1 billion characters from Wikipedia.\footnote{\url{http://mattmahoney.net/dc/enwik9.zip}}
For lexical information, we also extract the leftmost (in the order of the aligned words) child (\emph{c}), leftmost parent (\emph{p}) and leftmost grandchild (\emph{cc}). Leftmost and rightmost items are common features for transition-based parsers \cite{zhang2011transition,manningfast} but we found only leftmost to be helpful in our case.
All POS tags, dependencies and named entities are generated using Stanford CoreNLP \cite{corenlp}.
The accuracy of this classifier on the development set is 84\%.

%

Similarly, we train a binary classifier for deciding whether or not to create a reentrant edge after a \treduce{}: in this case we use word and POS embeddings for the two nodes being connected and their parent as well as dependency label embeddings for the arcs between them.

\subsection{Concept Identification}
\label{se & t2s}
This routine is called every time the transition classifier
decides to do a \tshift{};  it is denoted by $a(\cdot)$ in \S\ref{se & ts}. This component could be learned in a supervised
manner, but we were not able to improve on a simple heuristic, which works as follows: during training, for each \tshift{} decided by the oracle, we store the pair $(\beta_0, \overleftarrow{\conceptfunc}(i))$ in a phrase-table. During parsing, the most frequent graph $H$ for the given token is then chosen.  
In other words, $a(i)$ approximates $\overleftarrow{\conceptfunc}(i)$ 
by means of the graph most frequently seen among all occurrences of token $x_i$
in the training set. 



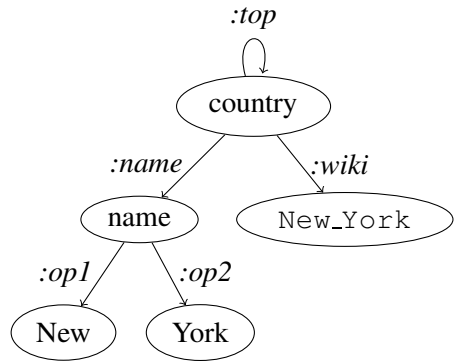
\begin{figure}[t]
 \centering
  \begin{tikzpicture}

        \draw (10,10) node(country)[ellipse,draw] {country};
        \draw (8.5,8.5) node(name)[ellipse,draw] {name};
        \draw (11.2,8.5) node(ny)[ellipse,draw] {{\tt New\_York}};
        \draw (7.5,7.0) node(new)[ellipse,draw] {New};
        \draw (9.3,7.0) node(york)[ellipse,draw] {York};

        \draw [->] (country) -- node[left]{\emph{:name}} (name);
        \draw [->] (country) -- node[right]{\emph{:wiki}} (ny);
        \draw [->] (name) -- node[left]{\emph{:op1}} (new);
        \draw [->] (name) -- node[right]{\emph{:op2}} (york);

        \draw [->] (country) edge[loop above]node{\emph{:top}} (country);

  \end{tikzpicture}

 \caption{Subgraph for ``\emph{New York}.''}
 \label{fig:names}
\end{figure}

An obvious problem with the phrase-table approach is that it does not generalize to unseen words.
In addition, our heuristic relies on the fact that the mappings observed in the data are correct, which is not the case when the JAMR-generated alignments contain a mistake. In order to alleviate this problem we observe that there are classes of words such as named entities and numeric quantities that can be disambiguated in a deterministic manner. 
We therefore implement a set of ``hooks'' that are triggered by the named entity tag of the next token in the sentence. These hooks override the normal \tshift{} mechanism and apply a fixed rule instead. For instance, when we see the token \emph{New York} (the two tokens are collapsed in a single one at preprocessing) we generate the subgraph of Figure~\ref{fig:names} and push its root onto the stack.
Similar subgraphs are generated for all states, cities, countries and people. 
We also use hooks for ordinal numbers, percentages, money and dates. 

\subsection{Edge Labeling}
Edge labeling determines the labels for the edges being created. Every time the transition
classifier decides to take an \tlarc{} or \trarc{} operation, the edge labeler needs to decide on a label for it. There are
more than 100 possible labels such as \emph{:ARG0}, \emph{:ARG0-of}, \emph{:ARG1}, \emph{:location}, \emph{:time} and \emph{:polarity}. We use a feed-forward neural network similar to the one we
trained for the transition classier, with features shown in Table~\ref{tab:features_labels}.
The accuracy of this classifier on the development set is 77\%.

\begin{table}[t]
  \centering
  \begin{tabular}{|l|l|}
	\hline
	name & feature template \\
    \hline
    depth & $d(\sigma_0)$, $d(\sigma_1)$ \\
    children & $\#c(\sigma_0)$, $\#c(\sigma_1)$ \\
    parents & $\#p(\sigma_0)$, $\#p(\sigma_1)$ \\
    lexical & $w(\sigma_0)$, $w(\sigma_1)$,\\
   & $w(p(\sigma_0))$, $w(c(\sigma_0))$, $w(cc(\sigma_0))$,\\
   & $w(p(\sigma_1))$, $w(c(\sigma_1))$, $w(cc(\sigma_1))$\\
    POS & $s(\sigma_0)$, $s(\sigma_1)$\\
    entities & $e(\sigma_0)$, $e(\sigma_1)$\\
    dependency & $\ell(\sigma_0,\beta_0), \ell(\beta_0,\sigma_0)$\\
    \hline
  \end{tabular}
  \caption{Features used in edge labeling. See Table~\ref{tab:features} for a legend of symbols.}
  \label{tab:features_labels}
\end{table}

\begin{eacl}
We constrain the labels predicted by the neural network in order to satisfy requirements of AMR. For instance, the label \emph{:top} can only be applied when the node from which the edge starts is the special $\troot$ node. Other constraints are used for the \emph{:polarity} label and for edges attaching to numeric quantities.
\end{eacl}

\begin{arxiv}
\paragraph{Labeling Rules}
Sometimes the label predicted by the neural network is not a label that satisfies the requirements of AMR. For instance, the label \emph{:top} can only be applied when the node from which the edge starts is the special $\troot$ node. In order to avoid generating such erroneous labels, we use a set of rules, shown in Table~\ref{tab:rules}. These rules determine which labels are allowed for the newly created edge so that we only consider those during prediction. Also ARG roles cannot always be applied: each Propbank frame allows a limited number of arguments. For example, while \emph{add-01} and \emph{add-02} allow for \emph{:ARG1} and \emph{:ARG2} (and their inverse \emph{:ARG1-of} and \emph{:ARG2-of}), \emph{add-03} and \emph{add-04} only allow \emph{:ARG2} (and \emph{:ARG2-of}).

\begin{table}[t]
  \centering
{\small
  \begin{tabular}{|l|l|l|l|}
    \hline
    label & ex. & start & end \\
    \hline
    :top & Yes & $\troot{}$ & \\
    :polarity & Yes &  & - \\
    :mode & Yes & & \small{\tt inter.$|$}\\
	& & & {\tt expr.$|$imp.} \\
    :value & No & & ``{\tt \textbackslash w$+$}'' {\tt |[0-9]$^+$} \\
    :day & No & {\tt d-ent} & {\tt [1|2|$\cdots$|31]} \\
    :month & No & {\tt d-ent} & {\tt [1|2|$\cdots$|12]$^+$} \\
    :year & No & {\tt d-ent} & {\tt [0-9]$^+$} \\
    :decade & No & {\tt d-ent} & {\tt [0-9]$^+$}\\
    :century & No & {\tt d-ent} & {\tt [0-9]$^+$}\\
    :weekday & Yes & {\tt d-ent} & {\tt [monday|$\cdots$|}\\
	&	&	& {\tt sunday]}\\
    :quarter & No & {\tt d-ent} & {\tt [1|2|3|4]$^+$} \\
    :season & Yes & {\tt d-ent} & {\tt [winter|fall|} \\
& & & {\tt spring|summer]$^+$}\\
    :timezone & Yes & {\tt d-ent} & {\tt $[$A$-$Z$]^3$}\\
    \hline
  \end{tabular}
}
  \caption{Labeling rules: For each edge label, we provide regular expressions that must hold on the labels at the start node (start) and the end node (end) of the edge. Ex. indicates when the rule is exclusive, d-ent is the AMR concept \emph{date-entity}, inter. is the AMR constant \emph{interrogative}, expr. is the AMR constant \emph{expressive}, imp. is the AMR constant \emph{imperative}.}
  \label{tab:rules}
\end{table}
\end{arxiv}

\section{Fine-grained Evaluation}
\label{se & metr}

Until now, AMR parsers were evaluated using the Smatch score.%
\footnote{Since Smatch is an approximate randomized algorithm, decimal points in the results vary between different runs and are not reported.  
This approach was also taken by \newcite{wang} and others.}
Given the candidate graphs and the gold graphs in the form of AMR annotations, Smatch first tries to find the best alignments between the variable names for each pair of graphs and it then computes precision, recall and F1 of the concepts and relations.
We note that the Smatch score has two flaws: (1)
while AMR parsing involves a large number of subtasks, the Smatch score consists of a single number that does not assess the quality of each subtasks separately;
(2) the Smatch score weighs different types of errors in a way which is not necessarily useful for solving a specific NLP problem. For example, for a specific problem concept detection might be deemed more important than edge detection, or guessing the wrong sense for a concept might be considered less severe than guessing the wrong verb altogether.

Consider the two parses for the sentence \emph{Silvio Berlusconi gave Lucio Stanca his current role of modernizing Italy's bureaucracy} in Figure \ref{fig:2parses}. At the top, we show the output of a parser (\emph{Parse 1}) that is not able to deal with named entities. At the bottom, we show the output of a parser (\emph{Parse 2}) which, except for \emph{:name}, \emph{:op} and \emph{:wiki}, always uses the edge label \emph{:ARG0}. The Smatch scores for the two parses are 56 and 78 respectively. Both parses make obvious mistakes but the three named entity errors in \emph{Parse 1} are considered more important than the six wrong labels in \emph{Parse 2}. However, without further analysis, it is not advisable to conclude that \emph{Parse 2} is better than \emph{Parse 1}.
In order to better understand the limitations of the different parsers, find their strengths and gain insight in which downstream tasks they may be helpful, we compute a set of metrics on the test set.
\begin{figure}
\tiny{
\begin{lstlisting}[frame=single]
((*@\textbf{g}@*) / (*@\textit{give-01}@*)
      (*@\textit{:ARG0}@*) ((*@\textbf{p3}@*) / (*@\textit{silvio}@*) (*@\textit{:mod}@*) ((*@\textbf{n4}@*) / (*@\textit{berlusconi}@*)))
      (*@\textit{:ARG1}@*) ((*@\textbf{r}@*) / (*@\textit{role}@*)
            (*@\textit{:time}@*) ((*@\textbf{c2}@*) / (*@\textit{current}@*))
            (*@\textit{:mod}@*) ((*@\textbf{m}@*) / (*@\textit{modernize-01}@*)
                  (*@\textit{:ARG0}@*) (*@\textbf{p4}@*)
                  (*@\textit{:ARG1}@*) ((*@\textbf{b}@*) / (*@\textit{bureaucracy}@*) (*@\textit{:part-of}@*) ((*@\textbf{c3}@*) / (*@\textit{italy}@*))))
            (*@\textit{:poss}@*) (*@\textbf{p4}@*))
      (*@\textit{:ARG2}@*) ((*@\textbf{p4}@*) / (*@\textit{person}@*) (*@\textit{lucio}@*) (*@\textit{:mod}@*) (*@\textit{stanca}@*)))

\end{lstlisting}
\begin{lstlisting}[frame=single]
((*@\textbf{g}@*) / (*@\textit{give-01}@*)
      (*@\textit{:ARG0}@*) ((*@\textbf{p3}@*) / (*@\textit{person}@*) (*@\textit{:wiki}@*) "Silvio_Berlusconi"
            (*@\textit{:name}@*) ((*@\textbf{n4}@*) / (*@\textit{name}@*) (*@\textit{:op1}@*) "Silvio" (*@\textit{:op2}@*) "Berlusconi"))
      (*@\textit{:ARG0}@*) ((*@\textbf{r}@*) / (*@\textit{role}@*)
            (*@\textit{:ARG0}@*) ((*@\textbf{c2}@*) / (*@\textit{current}@*))
            (*@\textit{:ARG0}@*) ((*@\textbf{m}@*) / (*@\textit{modernize-01}@*)
                  (*@\textit{:ARG0}@*) (*@\textbf{p4}@*)
                  (*@\textit{:ARG0}@*) ((*@\textbf{b}@*) / (*@\textit{bureaucracy}@*)
                        (*@\textit{:ARG0}@*) ((*@\textbf{c3}@*) / (*@\textit{country}@*) (*@\textit{:wiki}@*) "Italy"
                              (*@\textit{:name}@*) ((*@\textbf{n6}@*) / (*@\textit{name}@*) (*@\textit{:op1}@*) "Italy"))))
            (*@\textit{:ARG0}@*) (*@\textbf{p4}@*))
      (*@\textit{:ARG0}@*) ((*@\textbf{p4}@*) / (*@\textit{person}@*) (*@\textit{:wiki}@*) -
            (*@\textit{:name}@*) ((*@\textbf{n5}@*) / (*@\textit{name}@*) (*@\textit{:op1}@*) "Lucio" (*@\textit{:op2}@*) "Stanca")))
\end{lstlisting}
\caption{Two parses for the sentence ``\emph{Silvio Berlusconi gave Lucio Stanca his current role of modernizing Italy's bureaucracy}.''}
\label{fig:2parses}
}
\end{figure}

\textbf{Unlabeled} is the Smatch score computed on the predicted graphs after removing all edge labels. In this way, we only assess the node labels and the graph topology, 
which may be enough to benefit several NLP tasks because it identifies basic predicate-argument structure. For instance, we may be interested in knowing whether two events or entities are related to each other, while not being concerned with the precise type of relation holding between them.

\textbf{No WSD} gives a score that does not take into account word sense disambiguation errors.
By ignoring the sense specified by the Propbank frame used (e.g., \emph{duck-01} vs \emph{duck-02}) we have a score that does not take
into account this additional complexity in the parsing procedure. To compute this score, we simply strip off the suffixes from all Propbank frames and calculate the Smatch score.

Following \newcite{sawai}, we also evaluate the parsers using the Smatch score on noun phrases only (\textbf{NP-only}), by extracting from the AMR dataset all noun phrases that do not include further NPs.

As we previously discussed, reentrancy is a very important characteristic of AMR graphs and it is not trivial to handle. We therefore implement a test for it (\textbf{Reentrancy}), where we compute the Smatch score only on reentrant edges.

Concept identification is another critical component of the parsing process and we therefore compute the F-score on the list of predicted concepts (\textbf{Concepts}) too. Identifying the correct concepts is fundamental: if a concept is not identified, it will not be possible to retrieve any edge involving that concept, with likely significant consequences on accuracy. This metric is therefore quite important to score highly on.

Similarly to our score for concepts, we further compute an F-score on the named entities (\textbf{Named Ent.}) and wiki roles for named entities (\textbf{Wikification}) that consider edges labeled with \emph{:name} and \emph{:wiki} respectively. These two metrics are strictly related to the concept score. However, since named entity recognition is the focus of dedicated research, we believe it is important to define a metric that specifically assesses this problem. Negation detection is another task which has received some attention. An F-score for this (\textbf{Negations}) is also defined, where we find all negated concepts by looking for the \emph{:polarity} role. The reason we can compute a simple F-score instead of using Smatch for these metrics is that there are no variable names involved.

Finally we compute the Smatch score on \emph{:ARG} edges only, in order to have a score for semantic role labeling (\textbf{SRL}), which is another extremely important subtask of AMR, as it is based on the identification of predicate-argument structures.

Using this evaluation suite we can evaluate AMRs on a wide range of metrics that can help us find strengths and weakness of each parser, hence speeding up the research in this area.
Table \ref{tab:2parses2} reports the scores for the two parses of Figure \ref{fig:2parses}, where we see that \emph{Parse 1} gets a high score for semantic role labeling while \emph{Parse 2} is optimal for named entity recognition. Moreover, we can make additional observations such as that \emph{Parse 2} is optimal with respect to unlabeled score and that \emph{Parse 1} recovers more reentrancies.

\begin{table}[t]
  \centering
  \begin{tabular}{|l|c|c|}
    \hline
    Metric & First parse & Second parse\\
    \hline
    Smatch & 56 & 78\\
    Unlabeled & 65 & 100\\
    No WSD & 56 & 78\\
    NP-only & 39 & 86\\
    Reentrancy & 69 & 46\\
    Concepts & 56 & 100\\
    Named Ent. & 0 & 100\\
    Wikification & 0 & 100\\
    Negations & 0 & 0\\
    SRL & 69 & 54\\
    \hline
  \end{tabular}
  \caption{Evaluation of the two parses in Figure~\ref{fig:2parses} with the proposed evaluation suite.}
  \label{tab:2parses2}
\end{table}
%


%


\section{Experiments}
\label{section:experiments}
\label{se & exp}
We compare our parser\footnote{Our parser is available at \url{https://github.com/mdtux89/amr-eager}, the evaluation suite at \url{https://github.com/mdtux89/amr-evaluation} and a demo at \url{http://cohort.inf.ed.ac.uk/amreager.html}}
 against two available parsers: JAMR \cite{carbonell2014discriminative} and
CAMR \cite{wang,wang2boosting}, using the LDC2015E86 dataset for evaluation. Both parsers are available online\footnote{JAMR: \url{https://github.com/jflanigan/jamr}, CAMR: \url{https://github.com/c-amr/camr}.} and were recently updated for SemEval-2016 Task 8 \cite{flanigan2016cmu,wang2016camr}. However, CAMR's SemEval system, which reports a Smatch score of 67, is not publicly available. CAMR has a quadratic worst-case complexity (although linear in practice). In JAMR, the concept identification step is quadratic and the relation identification step is $O(|V|^2 \log |V|)$, with $|V|$ being the set of nodes in the AMR graph.


\begin{table}[t]
  \centering
  \begin{tabular}{|l|c|c|c|c|}
    \hline
    Metric & J'14 & C'15 & J'16 & Ours\\
    \hline
    Smatch & 58 & 63 & \textbf{67} & 64\\
    Unlabeled & 61 & \textbf{69} & \textbf{69} & \textbf{69}\\
    No WSD & 58 & 64 & \textbf{68} & 65\\
    NP-only & 47 & 54 & \textbf{58} & 55\\
    Reentrancy & 38 & 41 & \textbf{42} & 41\\
    Concepts & 79 & 80 & \textbf{83} & \textbf{83}\\
    Named Ent. & 75 & 75 & 79 & \textbf{83}\\
    Wikification & 0 & 0 & \textbf{75} & 64\\
    Negations & 16 & 18 & 45 & \textbf{48}\\
    SRL & 55 & \textbf{60} & \textbf{60} & 56\\
    \hline
  \end{tabular}
  \caption{Results on test split of LDC2015E86 for JAMR, CAMR and our \amreager{}. J stands for JAMR and C for CAMR (followed by the year of publication). Best systems are in bold.}
  \label{tab:results}
\end{table}

Table~\ref{tab:results} shows the results obtained by the parsers on all metrics previously introduced. On Smatch, our system does not give state-of-the-art results.
However, we do obtain the best results for \emph{Unlabeled} and \emph{Concept} and outperform the other parses for \emph{Named Ent.} and \emph{Negations}. Our score of \emph{Reentrancy} is also close the best scoring system, which is particularly relevant given the importance of reentrancies in AMR. The use of the \treduce{} transition, which targets reentrancies caused by control verbs, is critical in order to achieve this result.

The relatively high results we obtain for the unlabeled case suggests that our parser has difficulty in labeling the arcs.
Our score for concept identification, which is on par with the best result from the other parsers, demonstrates that there is a relatively low level of token ambiguity.
State-of-the-art results for this problem can be obtained by choosing the most frequent
subgraph for a given token based on a phrase-table constructed from JAMR alignments on the training data.
The scores for named entities and wikification are heavily dependent on the hooks mentioned in~\S\ref{se & t2s}, which in turn relies on the named entity recognizer to make the correct predictions. In order to alleviate the problem of wrong automatic alignments with respect to polarity and better detect negation, we performed a post-processing step on the aligner output where we align the AMR constant \emph{-} (minus) with words bearing negative polarity such as \emph{not}, \emph{illegitimate} and \emph{asymmetry}.


Our experiments demonstrate that there is no parser for AMR yet that conclusively does better than all other parsers on all metrics.
Advantages of our parser are the worst-case linear complexity and the fact that is possible to perform incremental AMR parsing, which is both helpful for real-time applications and to investigate how meaning of English sentences can be built incrementally left-to-right.

\begin{arxiv}
\section{Related Work}
\label{se & relevant}

The first data-driven AMR parser is due to
\newcite{carbonell2014discriminative}. The problem is addressed in two separate stages: concept
identification and relation identification. They use a sequence labeling algorithm to identify
concepts and frame the relation prediction task as a constrained combinatorial optimization problem.
\newcite{werling2015robust} notice that the difficult bit is the concept identification and propose a better way to handle that task: an action
classifier to generate concepts by applying predetermined actions. Other proposals involve a synchronous hyperedge replacement grammar solution \cite{peng2015synchronous}, a syntax-based machine translation approach \cite{pust2015using} where a grammar of string-to-tree rules is created after reducing AMR graphs to trees by removing all reentrancies, a CCG system that first parses sentences into lambda-calculus representations \cite{artzi2009broad}. A systematic translation from AMR to first order logic formulas, with a special treatment for quantification, reentrancy and negation, is discussed in \newcite{bos2016expressive}. In \newcite{microsoft}, a pre-existing logical form parser is used and the output is then converted into AMR graphs. Yet another solution is proposed by \newcite{searn_amr} who discuss a parser that uses SEARN \cite{searn}, a ``learning to search'' algorithm.

Transition-based algorithms for AMR parsing are compelling because traditional graph-based
techniques are computationally expensive. \newcite{wang} and \newcite{wang2boosting} propose a framework that parses a sentence
into its AMR structure through a two-stage process: a dependency tree is generated from the input sentence
through a transition-based parser and then another transition-based parser is used to generate the
AMR. The main benefit of this approach is that the dependency parser can be trained on a training
set much larger than the training set for the tree-to-graph algorithm. Others further built on this parser: \newcite{goodman2016noise} use imitation learning to alleviate the probem of error propagation in the greedy parser, while \newcite{barzdins2016riga} create a wrapper around it to fix frequent mistakes and investigate ensembles with a character level neural parser. More recently \newcite{emnlp2016} presented a non-greedy transition system for AMR parsing, based on \arcstandard{} \cite{nivre2004}.

AMR parsing as a whole is a complex task because it involves many subtasks including named
entity recognition, co-reference resolution and semantic role labeling. \newcite{sawai} do not
attempt at parsing AMR graphs for entire sentences but they instead handle simple noun phrases
(NPs).
They extract NPs from the AMR dataset only when they do not include further NPs, do not
include pronouns nor named entities.
Due to these restrictions, the AMRs are mostly trees and easier to handle than the original
AMR graphs. They approach this task using a transition based system inspired by \arcstandard{}.

AMR is not the only way to represent meaning in natural language sentences. Alternative semantic representations have been developed and studied, such as Boxer \cite{bos2004wide}, CCG \cite{surface_steedman,syntax_steedman} and UCCA \cite{abend2013universal}.
\end{arxiv}

\section{Conclusion}
\label{se:conclusion}
We presented a transition system that builds AMR graphs in linear time by processing the sentences left-to-right, trained with feed-forward neural networks. The parser demonstrates that it is possible to perform AMR parsing using techniques inspired by dependency parsing. 

We also noted that it is less informative to evaluate the entire parsing process with Smatch than to use a collection of metrics aimed at evaluating the various subproblems in the parsing process. We further showed that our left-to-right transition system is competitive with publicly available state-of-the-art parsers. Although we do not outperform the best baseline in terms of Smatch score, we show on par or better results for several of the metrics proposed. We hope that moving away from a single-metric evaluation will further speed up progress in AMR parsing.



\section*{Acknowledgments}

The authors would like to thank the three anonymous reviewers and Sameer Bansal, Jeff Flanigan, Sorcha Gilroy, Adam Lopez, Nikos Papasarantopoulos, Nathan Schneider, Mark Steedman, Sam Thomson, Clara Vania and Chuan Wang
for their help and comments.  This research was supported by a grant from Bloomberg
and by the H2020 project SUMMA, under grant agreement 688139.

\bibliography{library}
\bibliographystyle{eacl2017}

\end{document}